\documentclass[lettersize,journal]{IEEEtran}
\usepackage{amsmath,amsfonts}
\usepackage{algorithmic}
\usepackage{array}
\usepackage[caption=false,font=normalsize,labelfont=sf,textfont=sf]{subfig}
\usepackage{textcomp}
\usepackage{stfloats}
\usepackage{verbatim}
\usepackage[dvipsnames]{xcolor}

\usepackage{graphicx}
\usepackage{url}
\usepackage{multirow}
\usepackage[normalem]{ulem}
\useunder{\uline}{\ul}{}
\usepackage[T1]{fontenc}

\def\BibTeX{{\rm B\kern-.05em{\sc i\kern-.025em b}\kern-.08em
    T\kern-.1667em\lower.7ex\hbox{E}\kern-.125emX}}
\usepackage{balance}
\begin{document}
	\renewcommand{\bfdefault}{b}

% switch between changes marked mode and publish mode
%\newcommand{\edit}[1]{{\textcolor{blue}{#1}}}
\newcommand{\edit}[1]{{\textcolor{black}{#1}}}

\title{Chroma Clues: Leveraging Color Statistics to Detect Synthetic Images}
\author{Lea Uhlenbrock, Davide Cozzolino, Christian Riess
\thanks{Work was supported by Deutsche Forschungsgemeinschaft (DFG, German Research Foundation) as 
	part of the Research and Training Group 2475 "Cybercrime and Forensic Computing" 
	(grant number 393541319/GRK2475/2-2024).}}

% switch between submit mode and arXiv mode
\markboth{Preprint. Currently under review.}
{Chroma Clues: Leveraging Color Statistics to Detect Synthetic Images}

\maketitle

\begin{abstract}
The evolution and dissemination of AI-synthesized images is occurring at an
unprecedented rate. Image generators are making rapid progress in their goal of
perfectly imitating natural images, which also challenges image forensics.

In this work, we exploit an underexplored cue in current generative models,
namely their weakness to imitate color statistics of natural images.
We first show that the LPIPS loss used for training image generators is less sensitive to chrominance than to luminance, which may lead to statistical discrepancies in the colors of synthetic images.
Building on this observation, we then introduce six hand-crafted color transformations and a method to learn a task-optimized color transform to statistically expose generated images.
These transformations can be used in various ways. First, we define color-sensitive
features at pixel-level or patch-level. A simple, interpretable classifier
achieves with these features an average generalization accuracy of 93.27\% and strong robustness against six types of post-processing. Second,
we demonstrate that the transformations exhibit characteristic visual noise
patterns in natural and synthetic image areas, which enables an intuitive
visual image evaluation. Third, we demonstrate that the transforms can enhance
color patterns in generated images for improved multiclass attribution.

\end{abstract}


\section{Introduction} 
AI-based image generation 
pushes the boundaries of creativity. New ways of
storytelling, art, and design are becoming accessible to a broad audience.
Images from popular generators such as Midjourney~\cite{midjourney}, Stable
Diffusion~\cite{sd-code} and Adobe Firefly~\cite{firefly} improve rapidly in
visual quality, and are widely disseminated in printed publications and on the
internet.
While most synthetic images are created for entertainment, illustration, or marketing purposes, they also offer an increasing potential for malicious use such as fraud or disinformation.
Thus, ensuring the authenticity of images and preserving trust in images is a
pressing task.

\begin{figure}[ht]
	\centering
	\includegraphics[width=\linewidth]{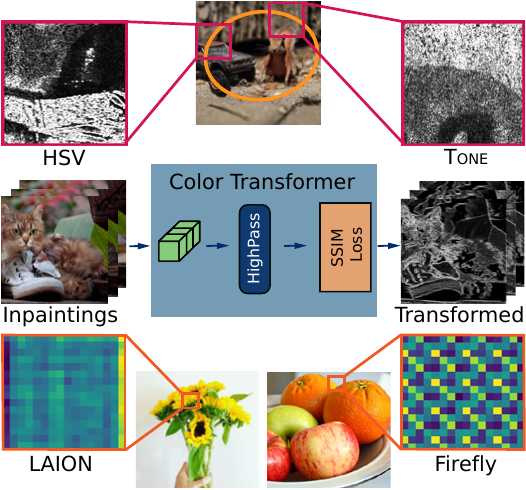}
	\caption{\textbf{Improved detectability of synthetic images through custom color transformations.}  \textbf{Top}: Noise residuals of an inpainting with HSV versus our color transform \textsc{TONE}. \textbf{Middle:} Visually optimized transformation obtained via CNN and perceptual loss. \textbf{Bottom}: \textsc{TONE} color fingerprints enable robust detection of LAION and Firefly images.}
	\label{fig:titelbild}
\end{figure}

Forensic research 
enables the blind detection of generated images. The toolbox for forensic analysis is diverse, adapting to different scenarios and challenges in the field.

Learning-based methods are highly effective and form the majority of approaches to detect synthetic images from Generative Adversarial Networks (GANs) and Diffusion Models (DMs). They use deep neural networks to identify subtle patterns and to differentiate between real and synthetic images. 
However, these approaches oftentimes require large quantities of training data.
Furthermore, the black box nature of such detectors may lack the type of interpretability that is required by law enforcement agencies or courts. 

Better interpretability is provided by physics-based cues, such as inconsistencies in the lighting
of generated images~\cite{farid2022lighting}.
However, their occurrence is typically constrained to specific scene content
and contextual circumstances. Moreover, one may argue that physical inconsistencies
oftentimes also affect the image quality, and may hence soon be weakened or
removed by the rapid progress in image generation.

Statistical traces form a third approach to detecting synthetic images.
Thorough analysis of synthetic images enables the careful crafting of features that distinguish them systematically from natural images. Such features may also be interpretable or generalizable across generators when traced back to certain systematic properties of generative models. 
Arguably the most widely known example are frequency fingerprints in GAN-generated images~\cite{marra2018do}. They can be traced back to the usage of specific upsampling architectures in the generator pipeline~\cite{durall2020watch, zhang2019detecting}. Hence, this property can enable the attribution of images to a specific generator architecture. 

In this work, we study statistical color traces, which may also be linked to an architectural element of neural networks. We show that the color formation in generative models may expose systematic traces that can be linked to the network loss. Existing literature contains some indications that generative models create color statistics that divert from natural images, but these observations have so far been reported in isolated instances and have not been investigated in depth.
In particular, several works report that detection of generated images in other color spaces, such as HSV and YCbCr, can significantly improve detection performance~\cite{zengDetectingDeepfakesAlternative, liIdentificationDeepNetwork2020a, moAIGeneratedFaceImage2022, qiaoCSCNetCrosscolorSpatial2023, aminExposingDeepfakeFrames, heDetectionFakeImages2019, chenRobustGANGeneratedFace2022}.
However, the \textit{origin} of color divergence between natural and synthetic images remains unexplored. We show that one plausible explanation for the effectivity of color statistics is their 
attribution to the perceptual loss for the training of generative models. As such, the divergence of color statistics can be understood as an architecture-dependent feature of image generators. We also explore hand-crafted and learned color transformations that maximize differences in color statistics (cf.~Fig.~\ref{fig:titelbild}).
A detector that operates on these transform spaces is interpretable and it generalizes well across generator models. We show that such color transforms can also improve inpainting localization (cf.~Fig.~\ref{fig:titelbild}) and model attribution\footnote{The code for this work will be made publicly available at \url{https://github.com/LeaUhlenbrock/ChromaClues}}.

\edit{Chrominance inconsistencies can provide useful cues for distinguishing real and synthetic images~\cite{uhlenbrock2024did}. 
Based on these findings, this work introduces a systematic framework for learning, analyzing, and exploiting color-based artifacts in image forensics. Our contributions are:}

	\begin{enumerate}
		\item \edit{We establish color transformations as a versatile representation for image forensics. To this end, we systematically explore input transformation strategies, to reveal a rich, previously underexplored feature space.}
		\item \edit{We formulate the discovery of effective and interpretable color transformations as an optimization problem, and introduce a corresponding CNN that learns such transformations directly from data.}
		\item \edit{We demonstrate that color descriptors can effectively tackle three core tasks in image forensics: detecting fully synthetic images, localizing manipulated regions such as inpaintings, and attributing images to their generative models.}
	\end{enumerate}

The remainder of this paper is organized as follows. Section~II reviews related work on synthetic image detection based on various features, in particular color clues. Section~III comprises an analysis of the perceptual loss used in AI-based image generators and its bias towards luminosity in comparison to chrominance. Section~IV provides a detailed description of the proposed method, including the manual crafting of alternative color transformations and an algorithm to optimize color transformations based on deep learning. Section~V presents the experimental setup and evaluation results. Finally, Section~VI concludes the paper and outlines future research directions.

\section{Related Work}
A significant body of research has been dedicated to the development of synthetic image detection techniques. 
There are multiple categories of traces that occur in synthetic images.

%\subsection{Visual Clues}
One class of traces are visual flaws caused by generators.
Early models of synthetic image generators often lacked a certain degree of comprehensive spatial understanding of a scene and thus often introduced small errors into the images like unusual amounts of fingers on a hand,
asymmetrical eye colors~\cite{maternExploitingVisualArtifacts2019}, unnatural color tones~\cite{leeClueCatcherCatchingDomainWise2023}, or broken and smudged structures~\cite{zhangPerceptualArtifactsLocalization2022}.
Further errors can be geometric inconsistencies such as reflections with incorrect perspective~\cite{faridPerspectiveConsistencyPaint2022}, or inconsistent scene lighting~\cite{farid2022lighting}.
Such inconsistencies can even be automatically detected, as demonstrated by Sarkar~\textit{et al.}~\cite{sarkar2024shadows}.
However, their manifestation depends on scene complexity~\cite{kamali2025characterizingphotorealismartifactsdiffusion}, limiting their overall reliability. Another inherent weakness of such traces is that they undermine visual quality, an essential objective for generative models, which leads to the assumption that they will gradually vanish from synthetic images. 
 
 More subtle clues of synthetic origin are frequency artifacts introduced by architectural peculiarities, such as the traces in the frequency domain that GAN models leave~\cite{marra2018do}.
  Similar traces have been shown to appear in DM-based images~\cite{corvi2023intriguing}. 
 While these artifacts provide a level of interpretability, in individual images they may be hard to detect, and weakened by post-processing.
 Moreover, Wesselkamp~\textit{et al.} show that they can easily be removed by frequency filtering and thus, classifiers relying on them can be fooled~\cite{wesselkamp2022misleading}.
 
 On a more local level, spatial relationships between image intensities display further peculiarities in comparison to natural images. Tan~\textit{et al.} demonstrate that upsampling used in generative models leaves more than frequency artifacts in GAN and DM-based images: The spatial inter-pixel relationship differs significantly from that of natural images~\cite{tan2024rethinking}. This can also be shown via the correlation between neighboring pixels~\cite{li2024improving,guarnera2020fighting}, the correlation between pixels in patches of low intra-variance~\cite{mallet2025simple} or the auto-correlation of the whole image displaying distinct patterns~\cite{corvi2023intriguing}.

Another line of research revolves around patterns in visual data that are not obvious to humans but can be learned by deep neural networks from large collections of training data. 
 Many works utilize such algorithms to automatically detect irregularities in pixel statistics of synthetic images~\cite{wang2020cnn, gragnaniello2021gan, liu2022detecting, tan2023learning, corvi2023detection, wang2023dire, heDetectionFakeImages2019, barniCNNDetectionGANGenerated2020,sha2023fake}. Zhong~\textit{et al.} utilize a deep neural network to show that the residuals of image patches can be used for detection~\cite{zhong2023patchcraft}. Chen~\textit{et al.} even demonstrate that this can be achieved using a single image patch~\cite{chen2024single}.
Recent approaches often focus on features from large pre-trained networks like CLIP~\cite{radford2021learning}. CLIP provides high-level feature representations which generalize well to various vision-related tasks.
They can effectively be used to distinguish real and synthetic images~\cite{ojha2023towards,lin2024robust,khan2024clipping,cozzolino2024raising,yan2024sanity} and even for source-attribution~\cite{moskowitz2024detecting,cioni2024clip} or inpainting detection~\cite{smeu2024declip}. 
While CLIP features have proven effective in detection tasks, there exists one concern about whether they are suited for long-term detection. 
CLIP features also are integrated into the training pipelines of generative models such as \mbox{DALL-E}~\cite{dall-e3} and are likely target to optimization strategies. During training, statistics of synthetic samples are pushed towards visual goals, including realism. Thus, all features that are subject to optimization of synthetic image generators may at some point converge towards those of natural images. While this is a purely theoretical scenario, the potential longterm-efficacy of forensic traces is an important property to consider.
Thus, we actively explore methods to extract traces from synthetic images that are orthogonal to generator optimization. One observation from the literature motivates a particular path forward: Synthetic images seem to exhibit systematic color peculiarities in comparison to natural images.
 \label{sec:colorcues}
 For example, McCloskey and Albright show that GAN-generated images are normalized during the generation process, which inherently limits the range of color intensities, leading to the absence of under- or overexposed areas in these images~\cite{8803661}.
 Barni~\textit{et al.} demonstrate that cross-channel co-occurences differ between GAN-generated and real images~\cite{barniCNNDetectionGANGenerated2020}.
 Multiple works further show that there exist statistical differences between real and synthetic images that become apparent when images are transformed into alternative colorspaces like YCbCr or HSV. Zeng~\textit{et al.} observe that generation artifacts are more visible in color spaces other than RGB and find that for detection, chrominance components of YCbCr and L*a*b* perform especially well~\cite{zengDetectingDeepfakesAlternative}. Li~\textit{et al.}~\cite{liIdentificationDeepNetwork2020a}, Mo~\textit{et al.}~\cite{moAIGeneratedFaceImage2022}, Qiao~\textit{et al.}~\cite{qiaoCSCNetCrosscolorSpatial2023}, Amin~\textit{et al.}, ~\cite{aminExposingDeepfakeFrames}, and He~\textit{et al.}~\cite{heDetectionFakeImages2019} utilize alternative color spaces to detect synthetic images and Chen~\textit{et al.}~\cite{chenRobustGANGeneratedFace2022} report that using the YCbCr colorspace can improve detector robustness. 
\edit{These works collectively suggest that there exist a mismatch between the color distributions between real and generated images.
In prior work~\cite{uhlenbrock2024did}, we showed that such a mismatch can arise from the perceptual optimization of generative models.
However, to the best of our knowledge, this insight has not been further explored.
In this work, we close this gap. We investigate color transformations as a general forensic representation framework. We hypothesize that there exists a broader manifold of color transformations capable of enhancing the detection of synthetic content. Hence, we explore a broader manifold of handcrafted transformations and introduce an optimization framework for learning forensic color representations directly from data. Further, we develop robust multi-scale feature extraction strategies resilient to diverse degradations, including AI-based compression and adversarial color manipulations.
Finally, we extend color-based forensic analysis beyond binary detection to multiclass attribution of images to their generating models.}

\section{Sensitivity Analysis of Generative Loss Functions to Color Information}

Divergences in the image formation processes of cameras and generative models give rise to consistent, systematic differences in how color information is encoded.
Natural images are discrete digital representations of real-world scenes, captured through physical interactions between light and a camera sensor. Pixel values in such images result from processes governed by the laws of optics and electronics. Consequently, the resulting color statistics are influenced by factors such as sensor characteristics, camera model, and scene content.
In contrast, synthetic images are generated by neural networks. These are designed to mimic natural images in the training data with respect to certain image statistics and an associated loss function. 
Rather than being constrained by physical laws, synthetic images are the result of this feature-based optimization. 

\subsection{Perceptual Optimization in Generative Models}
As generative models advance, they produce progressively fewer visual artifacts, presumably to improve user satisfaction and profitability. The vanishing of such traces constitutes a challenge for forensic analysis. Therefore, identifying traces that are unlikely to interfere with the perceptual optimization objectives of generative models may be key to developing robust, long-term strategies for synthetic image detection.

Transforming images from RGB to alternative color spaces, such as HSV or YCbCr, can enhance the detection performance of classifiers~\cite{zengDetectingDeepfakesAlternative,liIdentificationDeepNetwork2020a,moAIGeneratedFaceImage2022,qiaoCSCNetCrosscolorSpatial2023,aminExposingDeepfakeFrames,heDetectionFakeImages2019,chenRobustGANGeneratedFace2022}. This phenomenon can plausibly be attributed to the circumstance that image generators are primarily trained and optimized to synthesize images in the RGB color space. 
Consequently, their reproduction of color is designed to achieve visually plausible statistics specifically in RGB, rather than to reflect the complex, physically grounded relationships between colors in natural imagery.
\label{sec:inpaintings}
This limitation becomes particularly evident in hybrid images, such as inpaintings, which combine both real and synthetic regions. In Fig.~\ref{fig:titelbild}, we demonstrate this difference using an image modified by Midjourney~6. After converting the image to the HSV color space and applying a Laplacian high-pass filter to the Hue (H) channel, we extract residuals that highlight differences in color texture. The resulting image is linearly scaled for visibility. Even without further processing, the generated are exhibits a pattern that is visually different from the background. This observation supports the hypothesis that generative models do not faithfully replicate natural color relationships beyond the RGB space.
This example illustrates that optimization strategies used in generative models can have a significant impact on image properties. 

To identify such vulnerabilities in synthetic image generators, this work closely examines the architecture of the open-source Stable Diffusion implementation~\cite{sd-code}, with a particular focus on the optimization of its generation process. Stable Diffusion\textquoteright s image synthesis consists of two primary stages: first, the semantic generation of images within latent space, which includes the denoising steps common to all latent DMs; and second, the decoding stage, where images are translated from latent representations back into pixel space. We hypothesize that a statistical divergence in pixel-level distributions is introduced during this decoding phase. Stable Diffusion\textquoteright s decoder utilizes one specific module to evaluate perceptual quality of images in pixel space: the Learned Perceptual Image Similarity Patch Similarity (LPIPS)~\cite{zhang2018unreasonable}.

\begin{figure}[t]
	\centering
	\includegraphics[width=\linewidth]{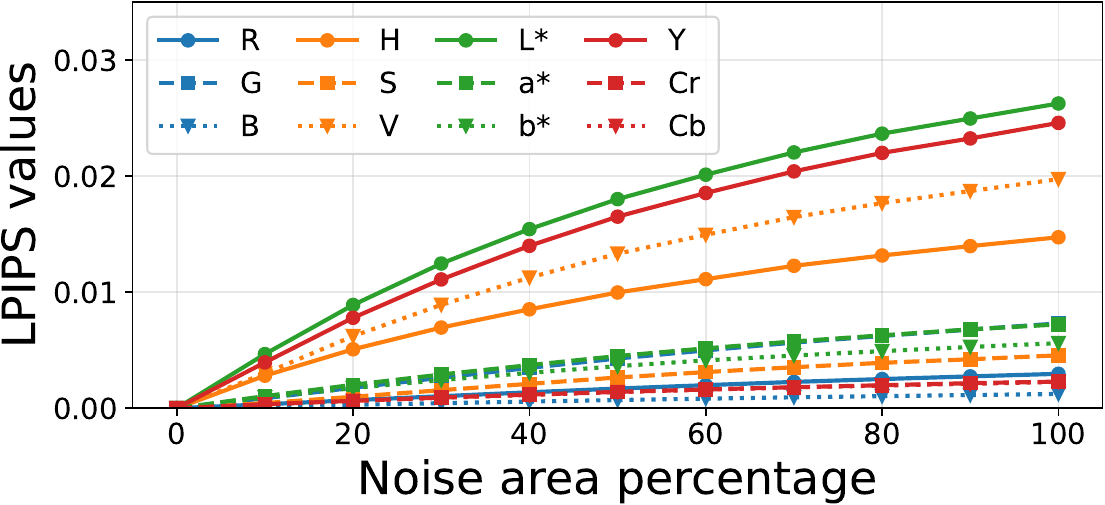}
	\caption{\textbf{Perceptual loss sensitivity across color spaces.} LPIPS consistently shows higher sensitivity to luminance than to chrominance components in all tested color spaces. }
	\label{fig:lpips}
\end{figure}

\subsection{Luminosity Bias in LPIPS Loss}

LPIPS computes the distance between a pair of images based on their features in the latent space of a pre-trained visual network. 
Zhang~\textit{et al.} have shown that these features are effective at imitating human visual perception~\cite{zhang2018unreasonable}.
Stable Diffusion uses this property to control the image quality during translation from latent to pixel space. This approach is a popular choice among a variety of generative networks~\cite{bruna2015super,dosovitskiy2016generating,johnson2016perceptual}. Although their architectural details are often not disclosed, it is reasonable to assume that many commercial generators also rely on similar mechanisms.
We design an experiment to investigate the sensitivity of LPIPS to color information, aiming to determine whether color is at least partially neglected during the optimization process.

Two copies of a natural image are used for the experiment. To one copy, Gaussian noise $\mathcal{N}(0,1)$ with mean $0$ and standard deviation $1$ is added to a certain percentage from $0-100$ of random image pixels, perturbing only one selected color channel at a time. This image is then converted back to RGB color space, keeping noise magnitudes of equivalent level. Then, the LPIPS metric between the unchanged image and the perturbed version is calculated to evaluate how perturbations in different color channels influence perceptual similarity. 
This procedure is repeated across the HSV, YCbCr, and L*a*b* color spaces, with the perturbed pixel area being increased incrementally. For reference, we also conduct this experiment in RGB space directly.

The results can be seen in Fig.~\ref{fig:lpips}. 
The $x$-axis indicates the percentage of pixels affected by the noise injection. The $y$-axis indicates the LPIPS loss.
Interestingly, perturbations in the luminance components have an effect on the perceptual distance that is around magnitude $4$ stronger than perturbances in the chrominance components. Specifically, additive noise in V (from HSV), Y (from YCbCr) and L (from L*a*b*) increases the LPIPS distance stronger than additive noise in the other color channels.
Hence, when such a perceptual metric is used in the training of generative neural networks to calculate the optimization loss, luminance irregularities are penalized much stronger than chrominance deviations.
Consequently, the chromaticities of synthetic images are more likely to exhibit statistical inaccuracies.
These inaccuracies can be revealed by color transformations, as demonstrated in Fig.~\ref{fig:titelbild}.

\section{Chrominance Noise Classification}
We assume that this observation is not limited to the particular choice of color spaces in Fig.~\ref{fig:lpips}. Instead, there may be a much larger subspace of transformations. 
In the following sections, we will investigate multiple approaches to craft alternative color transformations that reveal the traces of synthetic images. We show that they can be used to extract features that distinguish synthetic and real images and are able to enhance visual detectability of inpaintings. 

\subsection{Handcrafted Color Transformations}
It has been shown throughout the research on the detection of synthetic images that they display a wide range of forensic traces, reflecting the diversity of existing generator models. Therefore, most likely, one color transformation is hardly able to reveal all color-related peculiarities of such images. 
Fridrich~\textit{et al.} introduced a diverse set of spatial filters~\cite{fridrich2012rich} and later filters between color channels to capture fine-grained artifacts in images~\cite{goljan2014rich}. We build on the idea of extracting rich, discriminative features by developing a comprehensive set of handcrafted color transformations that expose diverse and complementary chromatic characteristics for improved forensic analysis.
One approach to constructing color transformations that are effective for detection of synthetic images is to analyze and reinterpret existing, well-performing transformations. Their underlying principles can be used for deriving other transformations.
For instance, the HSV color space reveals an instructive design: the components Hue (H) and Saturation (S) are computed by reordering the RGB channels per pixel based on their intensity and computing a relation between these rankings. Such mechanisms inspire new transformations that encode relational information between channels, rather than treating each channel independently. This is an effective way of revealing the inaccuracies of color relationships in synthetic images.
%We create more transformations that are logically related to the HSV definition. We further utilize the HSV, YCrCb and L*a*b* color space to create transforms that are based on the chrominance components of those color spaces. 
%This results in five custom transformations that are based on channel reordering and one transformation that combines chrominance components from HSV, YCrCb and L*a*b*. 
We craft one transformation based on the reordering of channels by pixel (\textsc{ORD}) and use it as the basis for four more: channel ratios (\textsc{RAT}), color tone (\textsc{TONE}), balance (\textsc{BAL}), and purity (\textsc{PUR}). Further, we use the chrominance components from HSV, YCbCr and L*a*b* color spaces and create three channels resembling saturation (\textsc{SAT}), similar to how chroma sometimes is calculated for color analysis tasks. The detailed formulas are listed in Tab.~\ref{tab:formula_table}.

%Each transformation based on the value-ordered channels is constructed in a way that yields values in the normalized range $[0,1]$. Further, w
Each transformation processes three input channels into three output channels. This structure allows for the capture of diverse chrominance characteristics within an image.

\begin{table}[t]
	\centering
	\renewcommand{\arraystretch}{1.8}

	\begin{tabular}{|l|l|}
		\hline
		
		\multicolumn{2}{|c|}{\textbf{ Transformations based on RGB}} \\
		\hline
		&$a = \max(R,G,B)$ \\
		\textsc{Value-ordered channels (Ord)}  &$b = \mathrm{median}(R,G,B)$   \\
		&$c = \min(R,G,B)$  \\
		\hline
		\textsc{Convolution-based (Conv)} &$\text{conv}(R,G,B)$  \\ \hline
		
		\multicolumn{2}{|c|}{\textbf{Transformations based on value-ordered channels}} \\
		\hline
		\textsc{Channel ratios (Rat)} & $\frac{b}{a},\frac{c}{b},\frac{c}{a}$ \\
		\hline
		\textsc{ Color Tones (Tone)}  &$\frac{b-c}{a-c},\frac{b-c}{a-c},\frac{a-b}{a-c}$ \\
		\hline
		\textsc{Color balance (Bal)} & $\frac{a-b}{a+b},\frac{a-c}{a+c},\frac{b-c}{b+c}$  \\ 
		\hline
		\multirow{2}{*}{\centering \textsc{Color purity (Pur)}} & $\mu = \frac{1}{3}(a+b+c)$ \\
		& $\frac{a}{\mu},\frac{b}{\mu},\frac{c}{\mu}$ \\
		\hline
		\multicolumn{2}{|c|}{\textbf{Transformations based on standard color spaces}} \\
		\hline
		\multirow{3}{*}{\centering \textsc{Saturation (Sat)}} &$\sqrt{Cb^2 + Cr^2}$ \\
		&$\sqrt{H^2 + S^2}$  \\
		&$\sqrt{a^{*2} + b^{*2}}$\\
		\hline
	\end{tabular}
	\vspace{2mm}
	\caption{\textbf{Custom color transformations for forensic analysis.} We group the defined transformations into RGB-based, value-ordered, and chrominance-derived categories, each emphasizing distinct color properties to support comprehensive forensic analysis.}
	\label{tab:formula_table}
	
\end{table}

%highpass=diagonal laplace

%\paragraph{additional way to craft transformations: learn them }

\subsection{Learned Color Transformations}
%\subsection{learned colortransformation}
Color transformations can also be learned if an objective function approximates the goal that synthetic and natural image areas shall be visually distinguishable.
Cozzolino and Verdoliva have already shown that training two Siamese Networks and using the Euclidean distance between residuals can be used to extract noise residuals that visually differ for images from different camera models~\cite{cozzolino2019noiseprint}. 
\edit{We hypothesize that this concept can be extended to systematically leverage a wide range of qualitative forensic observations for both improved automatic detection and their visual amplification by formulating optimization objectives that reflect these observations. In our setting, this corresponds to learning a color transformation, guided by an appropriate visual distance metric, that enhances the detectability of synthetic images. By maximizing the visual distance between natural and synthetic images under this metric, the transformation can be learned directly.}
 Several metrics exist for quantifying visual similarity between images, such as the previously discussed LPIPS. However, as we have shown, LPIPS exhibits limited sensitivity to chrominance information.
There exist multiple other quality metrics that can be used to optimize distinguishing visual image representations. To achieve a difference in residuals, a structure-focused metric such as the structural similarity index measure (SSIM) is a popular choice. It emphasizes local structural and luminance differences while maintaining sensitivity to chromatic variations. 

%\paragraph{we use SSIM to measure difference between transformed real/synth images}
%As discussed before, there exist many different calculations that can be used to achieve color transformations. 

Linear color transformations correspond to per-pixel mixing of the color channels. Hence, they can be modeled by a pixel-wise 
$1 \times 1 $ convolution. A composition of multiple such $1\times 1$
convolutions with non-linear activation functions allows the modeling of
non-linear color transformations, which are potentially even more expressive.
The parameters of these $1\times 1$ convolutions can be 
found via end-to-end training of a Convolutional Neural Network (CNN).

%\paragraph{we use SSIM in a contrastive loss setup}

The CNN consists of five sequential $1 \times 1$ convolutional layers, each with $3$ input and $3$ output channels. This leads to pixel-wise transformations independent of spatial context that enable the model to learn fine-grained color mappings while suppressing semantic variations.
Each layer is followed by batch normalization and a Leaky ReLU activation.
The network incorporates two residual connections to stabilize gradient flow and to preserve low-level chromatic features. Specifically, the network input is added element-wise to the input of the third convolutional layer and this intermediate result is subsequently added to the input of the fifth convolutional layer.
\label{sec:kernel}
The last layer of each network is initialized with a high-pass filter to extract noise residuals. We use a diagonal version of the Laplace high-pass filter $H$, applied per channel, with kernel coefficients $(0.25, 0, 0.25; 0,$$-1$$, 0; 0.25, 0, 0.25)$. This is inspired by the findings by Corvi~\textit{et al.} that the spectral energy of synthetic images compared to real images are more discriminative in diagonal direction~\cite{corvi2023intriguing}. 

We configure a contrastive SSIM-based loss.
First, we calculate the intra-class similarity for synthetic samples as
%\[
%\text{S}_{\text{rr}} = \frac{1}{N(N-1)} \sum_{i=1}^{N} \sum_{j=1, j \neq i}^{N} \text{sim}(r_i, r_j),
%\]
\begin{equation}
\text{S}_{\text{ff}} = \frac{1}{m(m-1)} \sum_{i=1}^{m} \sum_{\substack{j=1 \\ j \neq i}}^{m} \text{ssim}(f_i, f_j)
\label{eq:S_ff}
\end{equation}
and the inter-class similarity as
\begin{equation}
\text{S}_{\text{rf}} = \frac{1}{nm} \sum_{i=1}^{n} \sum_{j=1}^{m} \text{ssim}(r_i, f_j)
\end{equation}
where \(r_i, r_j\) are real samples and \(f_i,f_j\) fake samples both after applying the convolutional color-transformation and the high-pass filter, and \( \text{ssim}(\cdot, \cdot) \) is used to measure similarity between samples.

The training dataset (explained in more detail below) consists of sets of one real and three synthetic image patches that are semantically similar, which is why we omit the calculation of the Intra-Class-Similarity for real images. Instead, we minimize the contrastive loss: 
\begin{equation}
\mathcal{L} = \frac{\text{S}_{\text{rf}}}{\left( \text{S}_{\text{ff}} + \epsilon \right)}
\end{equation}
where \( \epsilon > 0 \) is a small constant to prevent division by zero.
 All convolutional layers are initialized using Kaiming initialization~\cite{he2015delving}. The model is implemented using PyTorch~\cite{pytorch} and trained for 35 epochs, using the validation loss as early stopping criterion. Adam is used for optimization, with a learning rate of $10^{-4}$ and weight decay factor of $10^{-5}$. The model is trained on a NVIDIA GeForce RTX 3090 GPU. 
 Example output of the Color Transformer network is shown in the middle row of Fig.~\ref{fig:titelbild}.

\label{sec:laplace}

In the Color Transformer, we want to capture structural differences and amplify subtle color cues into visually distinct patterns that are independent of high-level image content. Guillaro~\textit{et. al}~\cite{guillaro2025bias} demonstrate that semantic alignment between synthetic and real data samples allows detectors to learn subtle artifacts that differ between image sources. Based on this insight, we leverage the properties of the TGIF inpainting dataset~\cite{mareen2024tgif}.
\label{sec:tgif}
This dataset uses a set of real images to create inpaintings from Stable Diffusion~2 and SDXL. For each original image, three inpaintings are created per generator. Each of the images contains a rectangle of newly generated scene content, and the whole image is regenerated in the process. This regeneration leads to a maximum of semantic alignment between original images and inpaintings in the regions outside the bounding box. 
We use $2439$ original images,  $7320$ images from Stable Diffusion~2 and $7320$ images from SDXL in training. For validation, $341$ original images, $1023$ images from Stable Diffusion~2  and $1023$ images from SDXL are used. Each minibatch consists of one real image and its three associated inpainting variations from either Stable Diffusion~2 or SDXL.

\subsection{Feature Extraction}
The learned and hand-crafted color transformations are used to detect synthetic images from various sources. To this end, we leverage the color-transformed image using four different descriptors, an ensemble of classifiers and decision level fusion. 
We incorporate four levels of locality, using perceptual window sizes ranging from $16 \times 16$ to $1 \times 1$ pixels, which has shown to increase robustness in preliminary experiments.

\subsubsection{Color Fingerprints}
Synthetic image generators often leave periodic patterns in their output that can easily be detected by spatial averaging. This pattern can be seen as a fingerprint that can identify an images as synthetic.

We calculate this color pattern $\bar{P}$ using the following method:
Each image is color-transformed, and each of the three resulting single-channel images is divided into patches of size $16 \times 16$ pixels. 
Each patch $P_i$ is filtered with a diagonal Laplace kernel (cf. Sec.~\ref{sec:kernel}) per color channel, resulting in a filtered patch $\tilde{P}_i$.
The final color pattern $\bar{P}$ is obtained by averaging all high-pass filtered patches,

\begin{equation}
\bar{P} = \frac{1}{o} \sum_{i=1}^{o} H(P_i), \quad \text{for } i = 1, 2, \dots, o \enspace.
\end{equation}

\subsubsection{Patch Entropy}
Based on the observation that the color-transformed images show significantly different patterns between synthetic and real areas, we assume that synthetic image patches display subtle, systematic differences in the way color is organized on a small scale. 
Each image is color-transformed, yielding three single-
channel images. Each of the resulting single-channel images is divided into patches of $8 \times 8$ pixels. For each patch, the local Shannon entropy~\cite{shannon_entropy} is calculated. The entropy values are collected in a feature vector of size $4096$.

\subsubsection{Spatial Pixel Co-occurences}

High-resolution feature extraction follows a classical method, inspired by the Spatial Rich Models from Fridrich and Kodovský~\cite{fridrich2012rich}. This enables the capturing of pixel-neighborhood statistics that can distinguish synthetic and natural images.

Each image is color-transformed. The same diagonal Laplacian high-pass filter as before is applied to each to extract residuals. For each channel, spatial co-occurences are calculated as described below:
The residuals are uniformly quantized into $8$ discrete bins, spanning their mathematically defined minimum and maximum values. Then, local neighborhoods of three horizontally adjacent residuals are extracted across the image and accumulated into a global histogram to represent the co-occurrence distribution.

\subsubsection{Channel Co-occurences}
The method of highest locality, focused on a single pixel at a time, is based on the idea from Goljan~\textit{et al.}~\cite{goljan2014rich}. Their approach extends the work of Fridrich and Kodovský~\cite{fridrich2012rich} by collecting $1 \times 3$ neighborhood co-occurrences, but across color channels instead along spatial coordinates~\cite{goljan2014rich}. 

We adopt this idea and apply it to our images following a similar processing pipeline as in the previous features. Each image is color-transformed, after which the same high-pass filter as before is applied to each of the three resulting color channels. The filtered channels are then combined, and the channel-neighborhood co-occurrences are computed. This computation is analogous to the spatial co-occurrences, but instead of considering three spatial neighbors, the relationships among the three color-channel neighbors are used. The resulting values are quantized as described above into the $8$ bins, and then the co-occurences between these three channels are collected in a histogram.

\begin{figure}[tb]
	\centering
	\includegraphics[width=\linewidth]{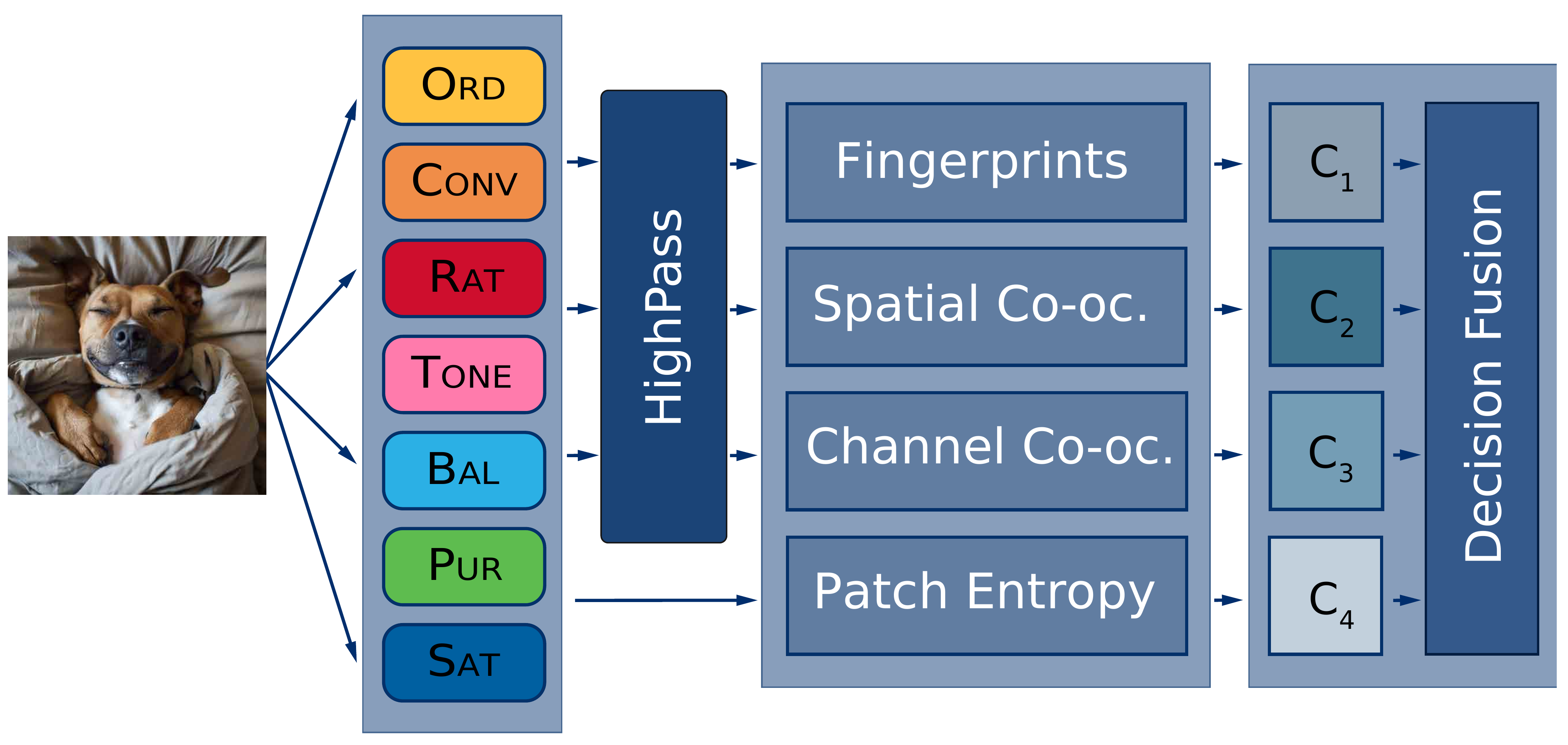}
	\caption{\textbf{Color Transformer pipeline overview.} Color-transformed images are filtered to extract noise residuals, which are classified by experts ($C_1$-$C_4$) trained on distinct descriptors. Their decisions are fused for final prediction.}
	\label{fig:pipelineall}
\end{figure}

\subsection{Classification through Color Ensembles}
A lightweight Support Vector Machine (SVM) classifier can be used to distinguish synthetic and natural images, given that the extracted features are highly discriminative. 
Such a lightweight classifier has various benefits:
It typically requires few training samples, improving practicality for synthetic image detection. Secondly, their simplicity makes these classifiers more interpretable and thus better suited for applications that require transparency in sensitive decision-making.
Consequently, we use lightweight SVM-based classification with a small amount of training samples to mimick real-life use cases.
To capture both fine-grained details and higher-level patterns, we construct an ensemble of four SVMs. Each model is trained on one of the four feature descriptors paired with a color transformation and uses an RBF kernel with a fixed random-state initialization. The final decision is achieved using soft voting on the Platt-scaled probabilistic scores~\cite{platt1999probabilistic} from the individual SVMs. 
In Fig.~\ref{fig:pipelineall}, the classification pipeline can be seen.

\section{Results}
The experimental evaluation is organized as follows: First, we describe the image data used for the experiments. Sec.~\ref{sec:preliminary} reports on the performance of the individual features and color transformations for synthetic image detection. Then, the ensemble of all features is compared to related works in Sec.~\ref{sec:comparison}. The experiments close with two additional use cases of the color features: The applicability of color fingerprints for multiclass detection and the usefulness of the color transformations to visually identify AI-generated image regions in inpaintings.

\begin{figure}[tb]
	\centering
	\includegraphics[width=\linewidth]{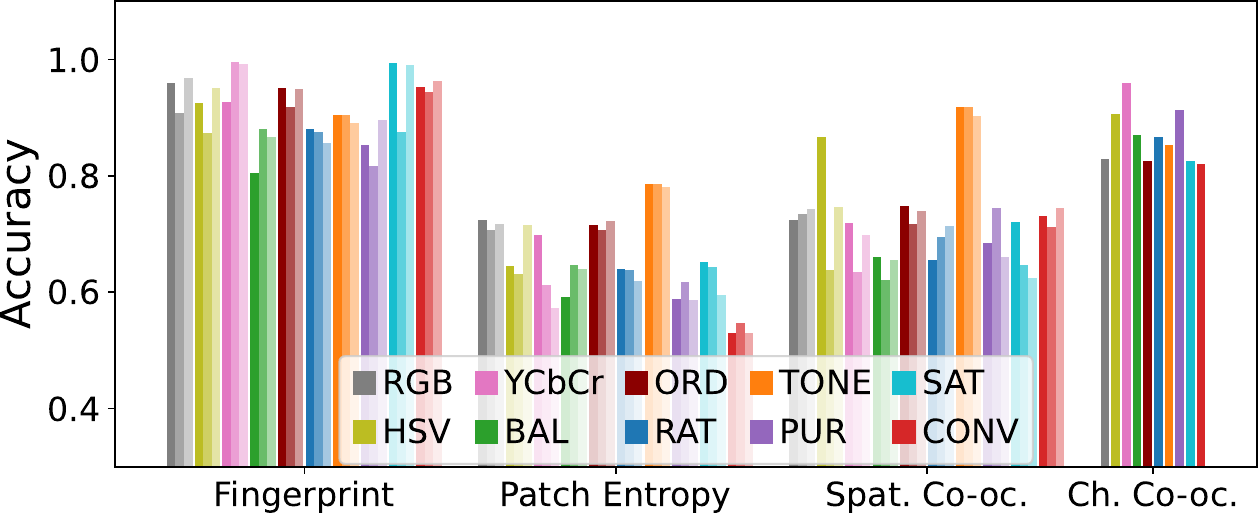}
	\caption{\textbf{Performance comparison of features and color transformations for detection.} Custom transformations such as \textsc{Tone} achieve accuracy comparable or better than pre-defined transformations.}
	\label{fig:colortransformperformance}
\end{figure}

\subsection{Evaluation of Individual Color Features}
\label{sec:preliminary}
To gain insight into how the basic components enhance classification of synthetic images, this experiment compares the detection accuracy of the individual classifiers. Each classifier is trained on a combination of one color transformation and feature extraction.
We use $400/100/400$ images for training/validation/testing from the COCO dataset and the same amount of images from Stable Diffusion.

The results are shown in Fig.~\ref{fig:colortransformperformance}. Fingerprints extracted from the first and third \textsc{Sat} channel outperform those from RGB, with accuracies of $99.4\%$ and $99.1\%$.
\edit{The \textsc{Tone} transformation outperforms all other color transformations when spatial neighborhood co-occurences are used as descriptors, with the highest-ranking channel features achieving an accuracy of $91.8\%$. All custom color transformations perform comparatively well across feature types, rendering them useful additions to pre-defined colorspaces for synthetic image detection. Notably, the level of locality significantly affects the performance ranking of color transformations, indicating that different transformations capture complementary information at different spatial scales. This demonstrates that locality and color transformation provide complementary cues, underscoring the benefit of combining both aspects.}

\subsection{Comparison with Related Work}
\label{sec:comparison}
In the following experiments, we compare our method against state-of-the-art works on synthetic image detection.
Robustness and generalizability are two important goals for detection. The proposed color ensemble can be tuned in either direction through the training protocol. We demonstrate this with two ensemble variants, which are denoted as \textit{Ours (Robust)} and \textit{Ours (Balanced)} in Tab.~\ref{tab:generalization} and Fig.~\ref{fig:robustness}. Both variants are described below.

\subsubsection{Datasets}
\label{sec:datasets}
The experiments are carried out on a dataset consisting of images by seven state-of-the-art DMs and two sources of real images.
The diffusion-based images are generated using popular models: \mbox{DALL-E}, Midjourney, Stable Diffusion and Adobe Firefly. We include legacy versions of the generators where possible to incorporate the evolution of synthetic images into our tests. Specifically, we use \mbox{DALL-E 2} (DE~2) data from Corvi et al.~\cite{corvi2023detection}, along with newly generated images from \mbox{DALL-E 3} (DE~3)~\cite{dall-e3}, Midjourney~5 (MJ~5)~\cite{midjourney}, Midjourney 6 (MJ~6)~\cite{midjourney}, Stable Diffusion~1.5 (SD)~\cite{sd-code}, Stable Diffusion XL (SDXL)~\cite{sdxl}, and Adobe Firefly (FF)~\cite{firefly}.
For generation of new images, we use captions from the COCO dataset~\cite{coco} as prompts. This ensures that the visual content of the diffusion-generated images aligns more closely with real-world images, encouraging the classifier to focus on low-level statistical features during detection instead of semantic differences.
Real images are from the COCO~\cite{coco} and the LAION~\cite{laion} datasets. The original LAION dataset includes product photographs consisting of mostly white pixels that are not relevant for our evaluation. Therefore, images that contain more than 20\% of white pixels are excluded, resulting in a more realistic image compilation. All images are center-cropped to $512 \times 512$ pixels.
Our models are trained on $400$ images from Stable Diffusion and $400$ images from COCO for training. The robust model uses $100$ additional validation images from both sources to determine the best classifier ensemble and fused decision threshold. Notably, the balanced model relies exclusively on hand-crafted color transformations; the learned Color Transformer appears only in the robust variant. Consequently, the balanced model provides a complementary evaluation setting that depends only on handcrafted color transformations, while the robust model utilizes the learned color-transformation.
The balanced model uses $100$ validation images from Midjourney 6 and $100$ validation images LAION on top for this. We use these same $1200$ images, including the Midjourney 6 and LAION validation images, to train those comparison works that are not available as pretrained models. 
Evaluation is conducted on datasets of size $800$: tests are performed on $400$ images from Stable Diffusion and $400$ images from COCO for the robustness tests. For each robustness test, all testing samples undergo one type of post-processing. The generalization performance is tested on $400$ unprocessed images of LAION and $400$ images of each of the diffusion models.

\subsubsection{Post-Processing}
To improve robustness of the individual SVMs, we use post-processed samples during training. We apply JPEG compression, Gaussian blur, Gaussian noise, and resizing as four standard post-processings.
\edit{We further include JPEG AI, a new compression standard that is expected to have a significant impact on image forensics as it becomes more widely adopted. Importantly, it can introduce artifacts similar to those of generative models~\cite{bergmann,cannas2024jpeg}, thereby creating a particularly challenging and practically relevant scenario for distinguishing natural from synthetic images.}
\label{sec:postprocessing}
We also add color post-processings that may potentially counteract the proposed color features.
In summary, we apply the following procedures:
\begin{itemize}
	\item \textbf{JPEG Compression}: The image is compressed using the JPEG standard with a quality factor \( q_{j} = 75 + 5k \), with \( 0 \le k \le 4 \).  
	\item \textbf{Gaussian Blur}: A spatial Gaussian blur is applied with \( \sigma_b \in \{1, 2, 3, 5, 8\} \).
	\item \textbf{Resizing}: The image is resized by a scaling factor \( r \in \{0.45, 0.75, 1.25, 1.55, 1.75\} \).  
	\item \textbf{Noise}: Gaussian noise \( N(\mu, \sigma_n^2) \) is added to each pixel, where \( \sigma_n \in \{5, 7, 9, 20, 50\} \).
	\item \textbf{JPEG AI Compression}: The image is compressed with the JPEG AI software using quality factors \( q_{a} \in \{75, 25, 6\} \).
	\item \textbf{Color attacks}: Histogram equalization, histogram stretching, contrast stretching and gamma correction are applied to the image.
\end{itemize}
We use OpenCV implementations of the procedures. For histogram equalization, we set the amount of bins to $100$, otherwise we use the default values. 
The total number of post-processing operations applied is $27$. Each operation is applied to $4$ random samples from the training set of each image source, yielding $108$ post-processed samples per source, i.e. $27\%$ of the training set. 

The threshold for soft voting and the color transformations used for the ensemble are optimized with respect to average accuracy on the validation samples.

\subsubsection{Methods for Comparison}
We compare our results against ten state-of-the-art works on synthetic image detection, including three methods utilizing the recently popular CLIP models and three works focused on color cues. 
Wang~\textit{et al.}~\cite{wang2020cnn}, Gragnaniello~\textit{et al.}~\cite{gragnaniello2021gan} and Tan~\textit{et al.}~\cite{tan2023learning} employ ResNet-50 architectures, while Corvi~\textit{et al.}~\cite{corvi2023detection} train their model on data generated via latent diffusion. Ojha~\textit{et al.}~\cite{ojha2023towards}, Khan~\textit{et al.}~\cite{khan2024clipping}, and Cozzolino~\textit{et al.}~\cite{cozzolino2024raising} leverage CLIP-based features for synthetic image detection. For evaluation of their methods, we utilize available pre-trained models.
Li~\textit{et al.} combine various color components and train an ensemble of Linear Discriminant Analysis classifiers~\cite{liIdentificationDeepNetwork2020a}. He~\textit{et al.} develop a shallow CNN to extract discriminative features, which are classified using a random forest~\cite{heDetectionFakeImages2019}. Barni~\textit{et al.} focus on cross-channel co-occurrence matrices of pixel values as discriminative features for detecting synthetic content~\cite{barniCNNDetectionGANGenerated2020}.

\begin{figure}[tb]
	\centering
	\subfloat[\label{fig:resize}]{%
		\includegraphics[width=0.45\linewidth]{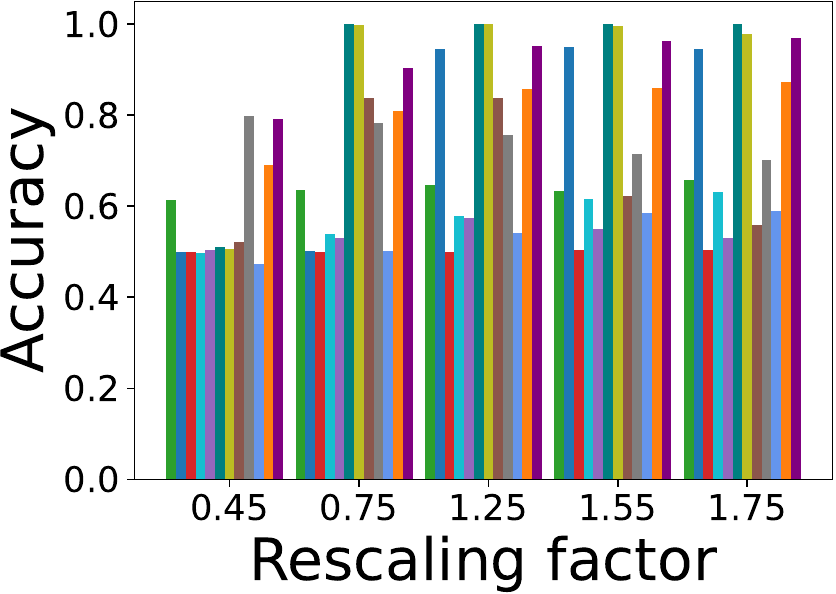}
	}
	\hfill
	\subfloat[\label{fig:blur}]{%
		\includegraphics[width=0.45\linewidth]{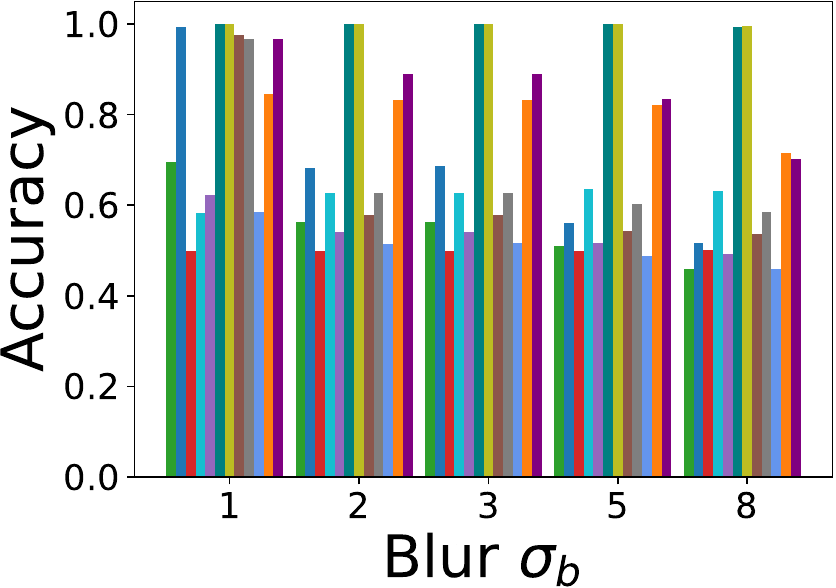}
	}
	
	\vspace{0.2em}
	
	\subfloat[\label{fig:jpg}]{%
		\includegraphics[width=0.45\linewidth]{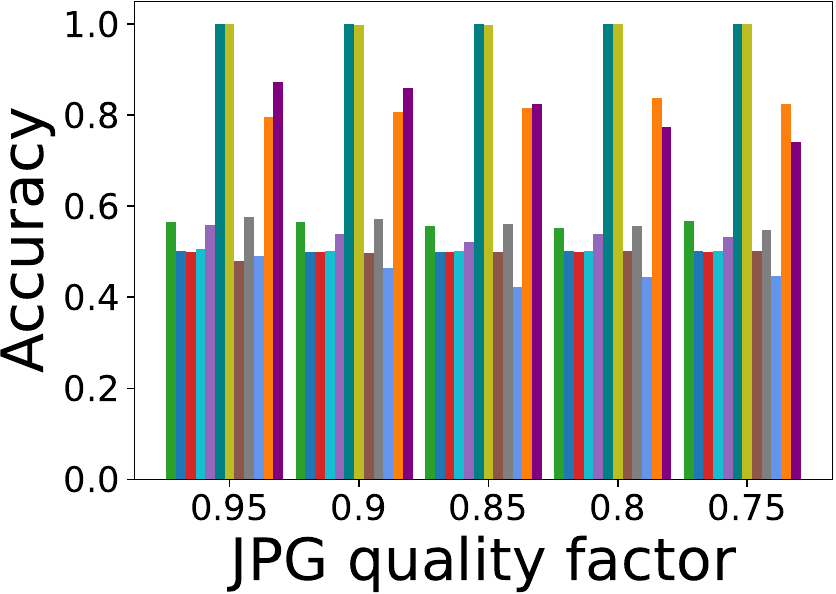}
	}
	\hfill
	\subfloat[\label{fig:noise}]{%
		\includegraphics[width=0.45\linewidth]{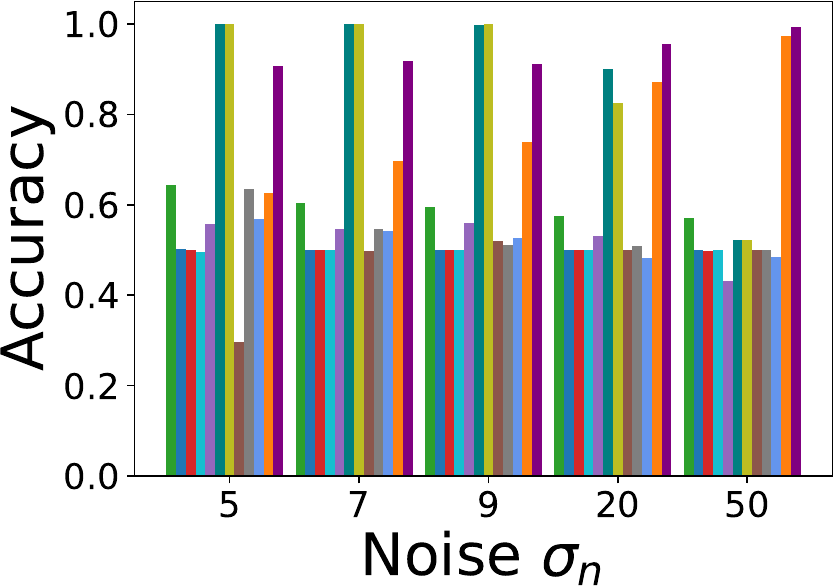}
	}
	
	\vspace{0.2em}
	
	\subfloat[\label{fig:colorattack}]{%
		\includegraphics[width=0.45\linewidth]{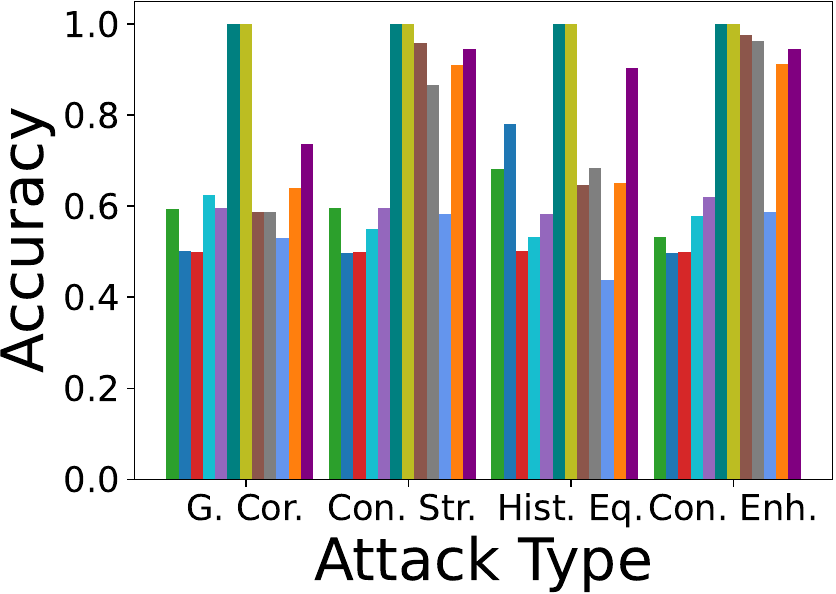}
	}
	\hfill
	\subfloat[\label{fig:ai}]{%
		\includegraphics[width=0.45\linewidth]{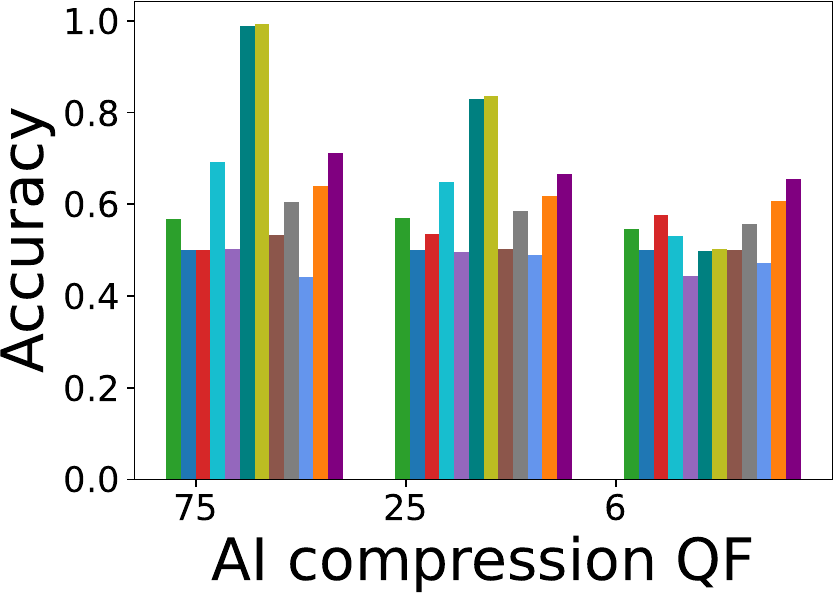}
	}
	
	\vspace{0.8em}
	\hfill
	\subfloat{%
		\includegraphics[width=0.95\linewidth]{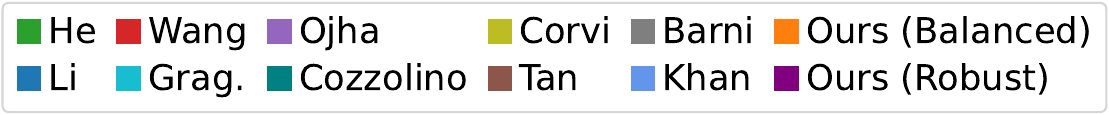}
	}

	\caption{\textbf{Robustness analysis of models under post-processing attacks.} Abbrevations: G.Cor. = Gamma Correction, Con. Str. = Contrast Stretching, Hist. Eq. = Histogram Equalization, Con. Enh. =  Contrast Enhancement}
	\label{fig:robustness}
\end{figure}

\subsection{Results on Robustness Experiments}
Fig.~\ref{fig:robustness} shows a comparison of our proposed models with the related works on post-processed images.
Remarkably, the robust and balanced models both surpass most other works under post-processing attacks. 
Cozzolino~\textit{et al.}~\cite{cozzolino2024raising} and Corvi~\textit{et al.}~\cite{corvi2023detection} perform better, but they use use 100k and 200k training images, respectively, whereas our model only uses $800/200$ training/validation images in total for the robust model.
The most degradation on feature quality is caused by gamma correction, strong blurring or JPEG compression and all levels of AI compression. Otherwise, the classifiers prove to be quite robust in this challenging setup.

\subsection{Results on Generalization Experiments}
\label{sec:generalization}
Table~\ref{tab:generalization} reports the results for the generalization experiments.
The balanced model outperforms related works by $>3.1\%$ accuracy on average. Further, detection accuracy of the tested individual generators never drops below $87.13\%$, even for \mbox{DALL-E 3} that is very prone to misclassification due to its forensic properties being highly similar to real images. Even the robust model achieves acceptable accuracy across datasets. 
Considering the light-weight classifier and high interpretability of the features, color features can be a reliable addition to synthetic image detection.

\begin{table*}[t]
	\begin{tabular}{l|>{\centering\arraybackslash}p{1.6cm}>{\centering\arraybackslash}p{1.6cm}>{\centering\arraybackslash}p{1.6cm}>{\centering\arraybackslash}p{1.7cm}>{\centering\arraybackslash}p{1.7cm}>{\centering\arraybackslash}p{1.6cm}>{\centering\arraybackslash}p{1.6cm}>{\centering\arraybackslash}p{0.45cm}}
		\hline
		\textbf{Method} & \textbf{DE 2} & \textbf{DE 3} & \textbf{FF} & \textbf{MJ 5} & \textbf{MJ 6} & \textbf{SD 1.5} & \textbf{SDXL} & \textbf{Avg} \\
		\hline
		
		Khan~\textit{et al.}~\cite{khan2024clipping}  & 83.75&	45.88&	84.25&	54.75&	53.38&	60.63&	68.75&	64.48 \\
		He~\textit{et al.}~\cite{heDetectionFakeImages2019}                       & 73.43&	53.29&	58.57&	58.14&	44.57&	62.00&	65.14	&59.31\\
		Li~\textit{et al.}~\cite{liIdentificationDeepNetwork2020a}                       & 33.38&	33.50	&58.13	&37.63&	34.50	&44.75&	33.38&	39.32\\
		Barni~\textit{et al.}~\cite{barniCNNDetectionGANGenerated2020}                    & 95.63&	48.63&	\underline{97.75}&	\underline{97.63}&	\underline{97.00}	&96.63	&97.38&	90.09 \\
		Ojha~\textit{et al.}~\cite{ojha2023towards}         & 78.88	&48.00&	78.25	&49.50	&49.25	&60.50	&64.25	&61.23 \\
		Cozzolino~\textit{et al.}~\cite{cozzolino2024raising}      &  48.50	&96.25&	69.37&	96.88	&96.88	&97.88	&97.88	&86.23\\
		Corvi~\textit{et al.}~\cite{corvi2023detection}       & 48.38	&\underline{97.75}	&58.25	&95.87	&96.63&	\underline{98.12}	&\underline{98.00}	&84.71 \\
		Wang~\textit{et al.}~\cite{wang2020cnn}        &52.62&	49.75&	55.25&	50.50	&50.00	&49.87&	49.87&	51.12  \\
		Tan~\textit{et al.}~\cite{tan2023learning}           &  \underline{95.75}&	48.37	&97.50	&95.12	&91.63	&97.38	&97.63	&89.05\\
		Grag.~\textit{et al.}~\cite{gragnaniello2021gan}       &63.00&	49.50&	91.87&	64.00	&51.62	&56.75	&60.62	&62.48  \\
		Ours (Robust)& 80.25&	68.25&	91.50&	80.63&	75.38	&94.75	&93.63	&83.48\\
		Ours (Balanced)          &         93.25&	87.13	&94.50	&94.50	&91.88	&96.00&	95.63&	\underline{93.27} \\
		\hline
	\end{tabular}
	\vspace{2mm}
	\caption{\textbf{Accuracy across synthetic image generators.} The best result per generator is underlined. While other works perform best in individual cases, our balanced model outperforms all compared works on average.}
	\label{tab:generalization}
\end{table*}

\begin{table*}[t]
	\begin{tabular}{l|>{\centering\arraybackslash}p{1.35cm}>{\centering\arraybackslash}p{1.35cm}>{\centering\arraybackslash}p{1.35cm}>{\centering\arraybackslash}p{1.35cm}>{\centering\arraybackslash}p{1.35cm}>{\centering\arraybackslash}p{1.35cm}>{\centering\arraybackslash}p{1.35cm}>{\centering\arraybackslash}p{1.35cm}>{\centering\arraybackslash}p{0.7cm}}
		\hline
		\textbf{Method} & \textbf{DE 2} & \textbf{DE 3} & \textbf{FF} & \textbf{MJ 5} &  
		\textbf{SD 1.3} & \textbf{SD 1.4} & \textbf{SD 2}  & \textbf{SDXL} & \textbf{Avg} \\
		\hline
		
		Khan~\textit{et al.}~\cite{khan2024clipping} & 84.63 & 41.13 & 85.38 & 58.63 & 71.38  & 72.00  & 74.25 & 79.50 & 70.86   \\
		He~\textit{et al.}~\cite{heDetectionFakeImages2019}                     & 61.29 & 44.29 & 56.57 & 49.86 & 59.43  & 57.57  & 42.00 & 52.00 & 52.88   \\
		Li~\textit{et al.}~\cite{liIdentificationDeepNetwork2020a}                                & 88.38 & 52.38 & 45.63 & 94.75 & 80.50  & 83.13  & 93.88 & 50.50 & 73.64   \\
		Barni~\textit{et al.}~\cite{barniCNNDetectionGANGenerated2020}                         & 94.63& 49.13 & 49.75 & 97.38 & 95.50  & 95.50  & 92.75 & 93.88 & \underline{83.56}   \\
		Ojha~\textit{et al.}~\cite{ojha2023towards}       & 82.40 & 46.30 & \underline{88.80} & 51.40 & 69.40  & 68.85  & 74.65 & 66.90 & 68.59  \\
		Cozzolino~\textit{et al.}~\cite{cozzolino2024raising}        & 48.10 & 47.80 & 77.95 & \underline{97.50} & \underline{97.80}  & 97.80  & \underline{97.35} & \underline{96.00} & 82.54  \\
		Corvi~\textit{et al.}~\cite{corvi2023detection}      & 48.30 & 48.10 & 57.15 & 97.15 & 98.10  & \underline{98.10}  & 96.65 & 95.30 & 79.86   \\
		Wang~\textit{et al.}~\cite{wang2020cnn}       & 52.90 & 49.75 & 55.95 & 50.25 & 50.10  & 50.15  & 51.70 & 51.75 & 51.57   \\
		Tan~\textit{et al.}~\cite{tan2023learning}          & \underline{96.65}& 53.55 & 50.85 & 81.35 & 97.10  & 97.65  & 76.80 & 98.40 & 81.54  \\
		Grag.~\textit{et al.}~\cite{gragnaniello2021gan}       & 62.85 & 49.40 & 68.20 & 52.70 & 57.05  & 57.55  & 56.95 & 61.90 & 58.32   \\
		Ours (Robust)          & 75.13	&48.38	&45.25	&76.50	&92.13&	92.50	&56.13	&92.13	&72.27 \\ 
		Ours (Balanced)        & 90.75	&\underline{61.75}	&53.38	&88.25	&94.88	&95.00	&76.00&	82.13&	80.27  \\  \hline
	\end{tabular}
	\vspace{2mm}
	\caption{\textbf{Accuracy across synthetic image generators of the Synthbuster Benchmark Dataset.} The best result per generator is underlined. While approaches based on deep feature extraction achieve higher absolute performance, our method maintains competitive accuracy across most generators under challenging generalization conditions.  }
	\label{tab:synthbuster}
\end{table*}

\subsection{Results on the Synthbuster Benchmark}
To contextualize our method within the broader landscape of deepfake detection approaches, we further evaluate it on a standard benchmark dataset.
Specifically, we employ the Synthbuster dataset with images from current generative models, namely
\mbox{DALL-E 2}, \mbox{DALL-E 3}, Adobe Firefly, Stable Diffusion~1.3 (SD 1.3), Stable Diffusion~1.4 (SD 1.4), Stable Diffusion~2 (SD 2), Stable Diffusion~XL and Midjourney 5~\cite{bammey2023synthbuster}.
All methods are evaluated on $400$ images of LAION and $400$ images from each
diffusion model from the Synthbuster test set.  These images are
generated from descriptions of RAISE images that were generated by Midjourney,
which leads to notable high-level content differences to the training data. 

Table~\ref{tab:synthbuster} summarizes the results. Despite the combined challenges of previously unseen generators and shifts in image content distributions, the average detection accuracy of the balanced model is $76.64\%$. The methods 
Barni~\textit{et al.}~\cite{barniCNNDetectionGANGenerated2020}, 
Cozzolino~\textit{et al.}~\cite{cozzolino2024raising}, and
Corvi~\textit{et al.}~\cite{corvi2023detection} achieve a higher average
accuracy, but their performance is more polarized to either very good
generalization performance on one generator or very poor generalization
performance on another. 
The proposed balanced model is for every detector better than guessing in these challenging generalization conditions.
Overall, this evaluation underscores the challenges posed by realistic generalization settings while demonstrating that the proposed method remains effective across a broad range of generators.

\begin{figure}[tb]
	\centering
	\includegraphics[width=\linewidth]{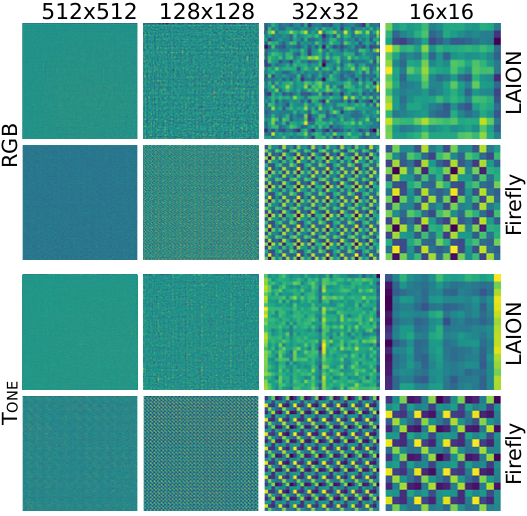}%coco: blockcomparison_dhp
	\caption{\textbf{Color fingerprints of Firefly and LAION.} Patches of sizes $512^2,128^2,32^2$, and $16^2$ averaged over $400$ images from Firefly and LAION from the blue Channel of RGB and the third \textsc{Tone} channel.}
	\label{fig:blocks}
\end{figure}
\begin{figure}[tb]
	\centering
	\includegraphics[width=\linewidth]{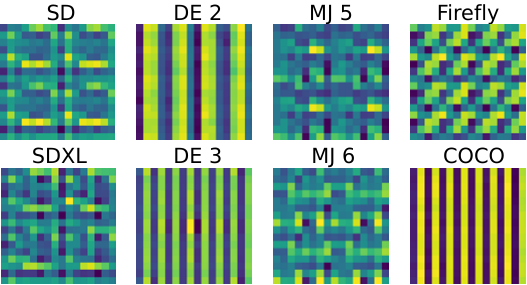}
	\caption{\textbf{Comparison of color fingerprints from \textsc{Sat}-transformed images.} Averaged first-channel fingerprints are shown for $400$ images from various synthetic models and COCO.}
	\label{fig:blockcomparison_all}
\end{figure}

\begin{figure}[tb]
	\centering
	\subfloat[RGB fingerprints\label{fig:multiclass_rgb}]{%
		\includegraphics[width=1\linewidth]{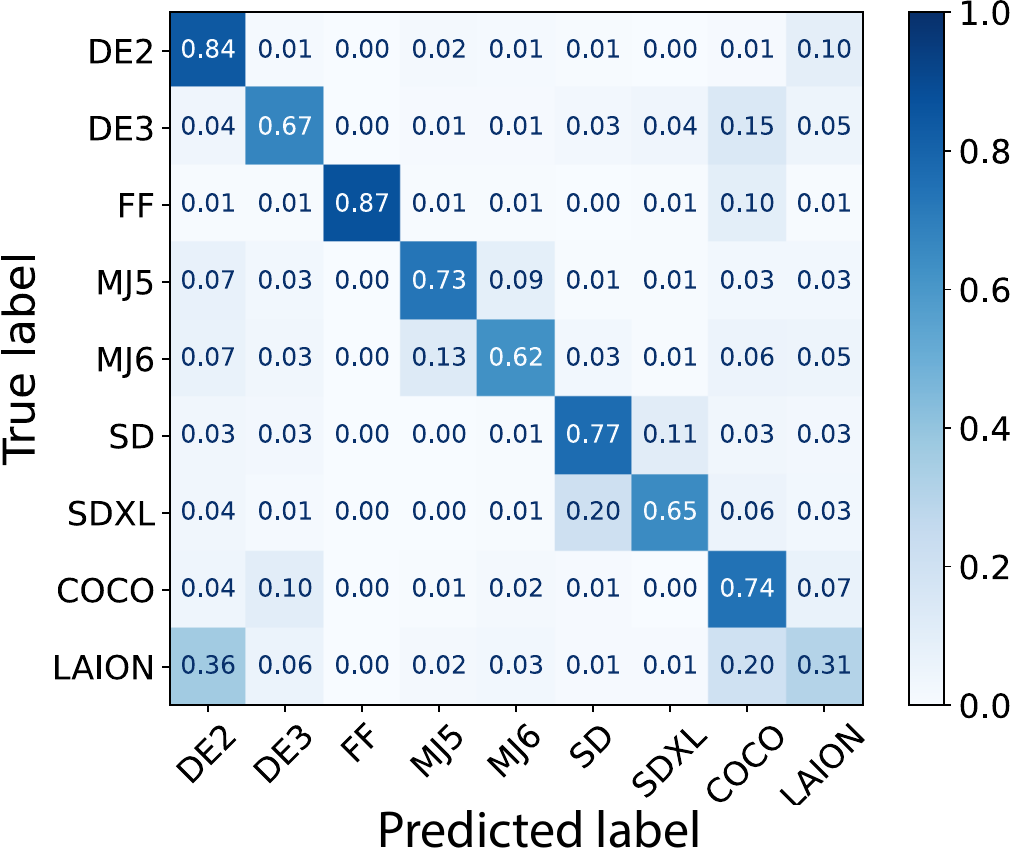}
	}
	\hfill
	\subfloat[\textsc{Sat} fingerprints \label{fig:multiclass_sat}]{%
		\includegraphics[width=1\linewidth]{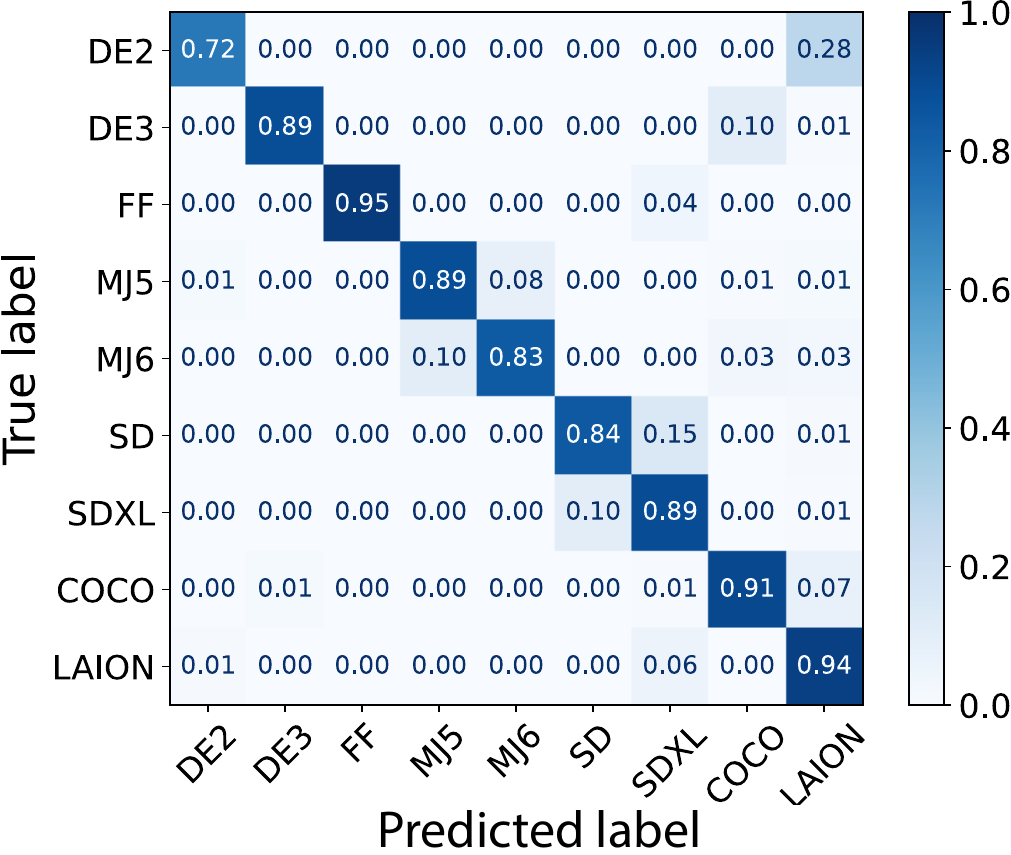}
		
	}
	
	\caption{\textbf{Source attribution confusion matrix.} \textsc{RGB} fingerprints (a) represent the baseline. \textsc{Sat} fingerprints (b) achieve considerably higher attribution accuracy. }
	\label{fig:multiclass}
\end{figure}

\subsection{Multiclass Detection with Color Fingerprints}
Another interesting effect seems to be influenced by the color transformations. They amplify periodic patterns hidden in synthetic images to increase their detectability. 
Marra~\textit{et al.} first showed that images from GANs leave fingerprints in the residuals of synthetic images~\cite{marra2018do}. Sinitsa~\textit{et al.} were able to adapt fingerprint extraction for diffusion-based images, using a deep network~\cite{sinitsa2024deep}. However, we demonstrate that custom color transformations are able to amplify the fingerprints in synthetic images without deep feature extraction. This not only strengthens established fingerprint-based discrimination but also improves interpretability by eliminating the need for deep feature extraction models.
Each high-passed and color-transformed image is split in patches of sizes $8 \times 8$ to $512 \times 512$ in powers of two. The resulting patches are averaged over $400$ images from the same source, resulting in a distinct color fingerprint.

Fig.~\ref{fig:blocks} shows these color fingerprints for LAION and Firefly, using the \textsc{Tone} and RGB color spaces and different patch sizes.
Interestingly, with the \textsc{Tone} color transformation, diagonal patterns appear in the averaged patches.
As also illustrated in Fig.~\ref{fig:blockcomparison_all} for the \textsc{Sat} color transformation, these artefacts exhibit significant variation across different image sources, thereby acquiring the characteristics of a fingerprint. 

Models from the same lineage tend to exhibit similar patterns, suggesting that shared backbones result in comparable fingerprints. While Stable Diffusion artifacts are oriented more horizontally and vertically, Midjourney and Firefly exhibit strong diagonal patterns. \mbox{DALL-E} on the other hand displays clearly visible vertical lines. It can be noted that the \mbox{DALL-E 3} API outputs JPEG images and the COCO images we used also were saved as JPEG. Due to the spatial alignment of DCT-blocks used in JPEG-compression with the blocks used for fingerprint extraction, such artifacts can be expected and would explain the similarity between \mbox{DALL-E 3} and COCO patterns.
 
\edit{Color fingerprints can also be used for source attribution. We extract the color patterns from the individual RGB channels and the best-performing color transformation from the experiment in \ref{sec:preliminary}, the \textsc{Sat} channels. We train an SVM with an RBF kernel on residuals from the $400$ training and $100$ validation dataset of each image source, totalling $4500$ training color patterns, and test on $400$ color patterns from the test-dataset from each source. Fig.~\ref{fig:multiclass} shows the confusion matrix for the best-performing RGB and \textsc{Sat} channels in comparison. We report the ratio of correctly as well as wrongly predicted images in our test. The \textsc{Sat} channel substantially improves source-attribution performance. For the RGB residuals, correct-prediction rates for the lower-performing classes typically fall between $30–-65$\%, whereas for \textsc{Sat} they increase to approximately $70-–85$\%. Likewise, the upper range of accuracies shifts from roughly $85$\% under RGB to well above $90$\% when using \textsc{Sat}. One outlier is \mbox{DALL-E 2}, which gets predicted as LAION more often when using \textsc{Sat}. However, the ratio of correctly predicted labels improves strongly for all other image sources. Another interesting effect is that the architectural similarity between Midjourney and Midjourney 6 and Stable Diffusion and SDXL is still recognizable using the \textsc{Sat} fingerprints, while other, more detrimental class confusions are strongly reduced in comparison to RGB. This illustrates that the color transformations offer additional application possibilities that are relevant for the identification of synthetic images.}

\begin{figure*}[tb]
	\centering
	\includegraphics[width=\linewidth]{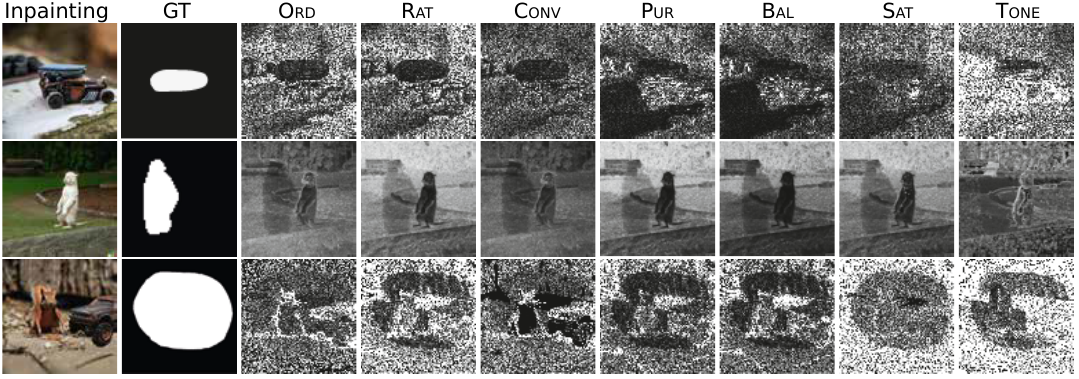}
	\caption{\textbf{Color transformations highlight differences in noise residuals.} The ground truth (GT) marks the edited region. Color transformations emphasize this area with varying precision, with different transformations performing best per image.}
	\label{fig:colortransforms}
\end{figure*}

\begin{figure}[tb]
	\centering
	\includegraphics[width=\linewidth]{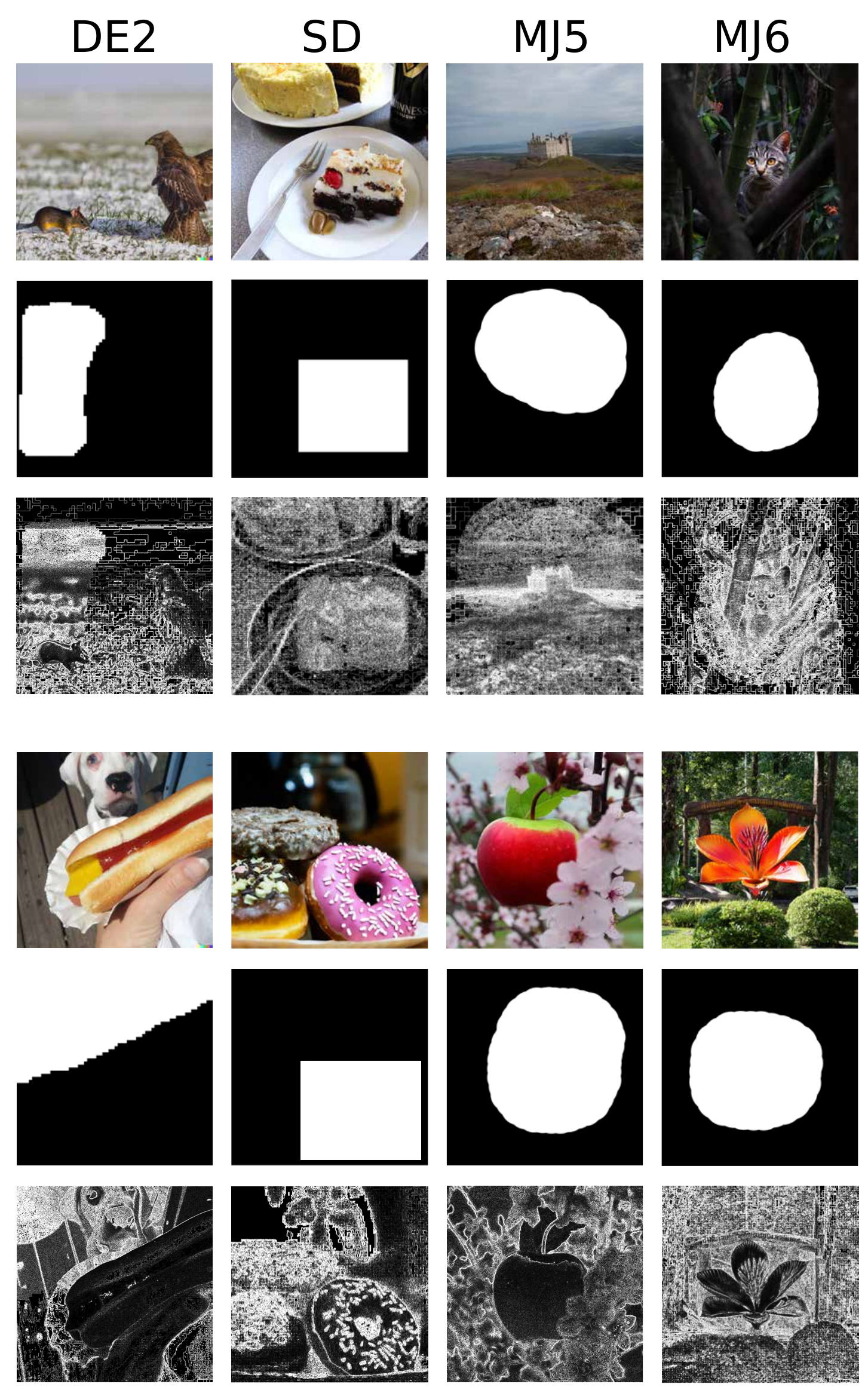}
	\caption{\textbf{High-pass filtered color-transformed images reveal inpainted regions.} Detection effectiveness varies with image content and is generally better in low-frequency areas.
	}
	\label{fig:successfails}
\end{figure}

\subsection{Qualitative Results}
We demonstrate that the color transformations crafted manually and trained via the CNN increase the visual detectability of inpaintings by repeating our experiment as seen in Fig.~\ref{fig:titelbild}. Multiple inpaintings are used, with our color transformations and a high-pass filter applied to them. Image intensities are scaled for better visibility. Fig.~\ref{fig:colortransforms} shows the results. The color transformations indeed provide visual detectability. 
They clearly highlight the synthetic part of the used images. We note that this effect is best visible when multiple transformations are applied and the resulting images are compared, as each transformation reveals slightly different irregularities in synthetic image residuals. 
For example, homogenous and textured areas react differently to the transformations. The \textsc{Sat} color transformation clearly defines the outline of the edited area in the bottom image in Fig.~\ref{fig:colortransforms} with highly different textures present, but it is less useful for the slightly more texture-poor, more color-diverse image in the top row. 
One question that arises is whether using an overlap of extracted noise residuals from different color transformations yields a visually improved result. We observe that the cumulative noise strength in this scenario overrides the distinctive differences between real and synthetic image parts in individual residual images. However, by comparing the residuals side-by-side or combining them in a more structured way, an analyst can reinforce consistent artifacts, suppress transformation-specific noise, and ultimately narrow down the region of potential manipulation more reliably than with any single residual alone.
% ich habe das auch bei schwierigen bildern beobachtet - irgendeine der transformationen revealed fast immer irgendwas! Nur eben nicht immer alle. Man muesste als analyst so bisschen an der helligkeit schrauben und die transforms vergleichen, und dann sieht man ganz oft was

We emphasize that this result is based entirely on simple transformations applied to the image. No black-box algorithm is used, with the exception on the training of the learned transformation with the Color Transformer. However, the Color Transformer is constrained to finding a simple color transformation due to its $1 \times 1$ convolutions operating on one single pixel at a time. The simple, transparent operations increase intuitiveness and accessibility.
Furthermore, reducing reliance on complex, opaque systems enhances the likelihood that forensic results will stand up to scrutiny in legal and investigative contexts.

\subsection{Limitations}
Fig.~\ref{fig:titelbild} shows that some color transformations amplify differences between synthetic and natural image noise residuals and thus significantly improve the visual detectability of inpaintings. However, this effect is not independent of image quality and content. While this phenomenon appears for inpaintings based on high-resolution images, artifacts of lower image quality can discrupt the fidelity of the resulting residual markings. In Fig.~\ref{fig:successfails}, we compare some sample inpaintings created using Stable Diffusion, \mbox{DALL-E 3}, and Midjourney~5 and 6. It can be seen that the visibility of the markings is higher wenn the border of the edited area is located in lower-frequency image areas. In images generated by Midjourney, a distinct double-radius marking often appears, suggesting that the software applies a blending or feathering technique to smooth the transition between the generated and original regions. This further complicates the precise boundary of the edited area. Nevertheless, the transition itself is oftentimes effectively revealed.
However, color transformations can often effectively reveal this transition.

\section{Conclusion}

\edit{Synthetic image generators such as Stable Diffusion use a loss in their decoder that optimizes visual quality of images. We demonstrate that the LPIPS loss as one such example prioritizes luminance over chrominance, resulting in subtle color artifacts in synthetic images. These artifacts become visually apparent in noise residuals when specific color transformations are applied, enabling effective detection. 
We demonstrate that color transformations form a rich and underexplored design space for image forensics by proposing six handcrafted transformations.
Furthermore, we formulate the search for color transformations as an optimization problem aimed at maximizing the visual divergence between synthetic and real images. This approach enables automated discovery of novel traces while maintaining high feature interpretability.
Our custom color transformations serve as the foundation to the extraction of distinctive features that generalize well across image generators. By incorporating multiple levels of spatial locality during feature extraction, our method ensures robust and effective detection of synthetic images. 
Beyond binary detection, we show that different transformations expose distinct forensic artifacts, enabling both qualitative analysis for inpainting localization and improved fingerprint-based multiclass attribution between generative models. Fingerprints, in particular, can improve interpretability and thereby also support emerging transparency and accountability requirements as outlined in the European AI Act~\cite{aiact}.}

In future work, it will be interesting to explore more methods of learning color transformations that enhance visual detectability of inpaintings. 
Furthermore, persuing qualitative robustness of the visible extracted color residuals along generator evolution and under post-processing would be valuable. 
Additionally, integrating textural information into color feature extraction may be a way to improve detection accuracy. For this, it would be valuable to analyze the strength and consistency of color traces in homogenous compared to highly textured regions. 
Further, a color-based analysis and classification of small image areas could be a promising direction towards automatically detecting inpaintings or marking image areas that reveal the synthetic origin of an image to an analyst.

\bibliographystyle{IEEEtran}
\bibliography{ref_opt}

\end{document}